\documentclass[nohyperref]{article}

\usepackage{microtype}
\usepackage{graphicx}
\usepackage{subfigure}
\usepackage{booktabs} 
\usepackage[frozencache,cachedir=mintedcache]{minted}

\usepackage{hyperref}

\usepackage[accepted]{icml2022}

\usepackage{amsmath}
\usepackage{amssymb}
\usepackage{mathtools}
\usepackage{amsthm}

\usepackage[capitalize,noabbrev]{cleveref}

\theoremstyle{plain}

\theoremstyle{definition}

\theoremstyle{remark}

\usepackage[textsize=tiny]{todonotes}

\usepackage{xspace}
\usepackage{units}

\newcommand{\mymath}[1]{\ensuremath{#1}\xspace}

\newcommand{\reals}{\mymath{\mathbb R}}

\definecolor{myred}{rgb}{0.8,0,0}
\definecolor{mygreen}{rgb}{0,0.6,0}
\definecolor{myblue}{rgb}{0,0,0.7}

\newcommand{\dvd}{{\sc d}v{\sc d}\xspace}

\newcommand{\qdpg}{{\sc qd-pg}\xspace}

\newcommand{\cpu}{{\sc cpu}\xspace} 
\newcommand{\ram}{{\sc ram}\xspace} 
\newcommand{\cpus}{{\sc cpu}s\xspace}

\newcommand{\tddd}{{\sc td3}\xspace} 
\newcommand{\sac}{{\sc sac}\xspace} 
\newcommand{\dqn}{{\sc DQN}\xspace} 
\newcommand{\cemrl}{{\sc cem-rl}\xspace} 
\newcommand{\pbt}{{\sc pbt}\xspace} 
\newcommand{\cem}{{\sc cem}\xspace} 
 
\newcommand{\pssstddd}{{\sc p3s-td3}\xspace}

\newcommand{\pgame}{{\sc pga-map-elites}\xspace}

\newcommand{\jit}{{\sc jit}\xspace}
\newcommand{\jax}{{\sc jax}\xspace}
\newcommand{\torch}{{\sc pytorch}\xspace}
\newcommand{\acme}{{\sc acme}\xspace}
\newcommand{\mlp}{{\sc mlp}\xspace}
\newcommand{\cuda}{{\sc cuda}\xspace}

\icmltitlerunning{Fast Population-Based Reinforcement Learning on a Single Machine}

\begin{document}

\twocolumn[
\icmltitle{Fast Population-Based Reinforcement Learning on a Single Machine}




\begin{icmlauthorlist}
\icmlauthor{Arthur Flajolet}{instadeep}
\icmlauthor{Claire Bizon Monroc}{instadeep}
\icmlauthor{Karim Beguir}{instadeep}
\icmlauthor{Thomas Pierrot}{instadeep}

\end{icmlauthorlist}

\icmlaffiliation{instadeep}{InstaDeep Ltd}

\icmlcorrespondingauthor{Arthur Flajolet}{a.flajolet@instadeep.com}
\icmlcorrespondingauthor{Thomas Pierrot}{t.pierrot@instadeep.com}


\vskip 0.3in
]



\printAffiliationsAndNotice{}  

\begin{abstract}
Training populations of agents has demonstrated great promise in Reinforcement Learning for stabilizing training, improving exploration and asymptotic performance, and generating a diverse set of solutions. However, population-based training is often not considered by practitioners as it is perceived to be either prohibitively slow (when implemented sequentially), or computationally expensive (if agents are trained in parallel on independent accelerators). In this work, we compare implementations and revisit previous studies to show that the judicious use of compilation and vectorization allows population-based training to be performed on a single machine with one accelerator with minimal overhead compared to training a single agent. We also show that, when provided with a few accelerators, our protocols extend to large population sizes for applications such as hyperparameter tuning. We hope that this work and the public release of our code will encourage practitioners to use population-based learning more frequently for their research and applications.
\end{abstract}

\section{Introduction}
\label{sec:introduction}

\begin{figure}
    \centering
    \includegraphics[width=0.45\textwidth]{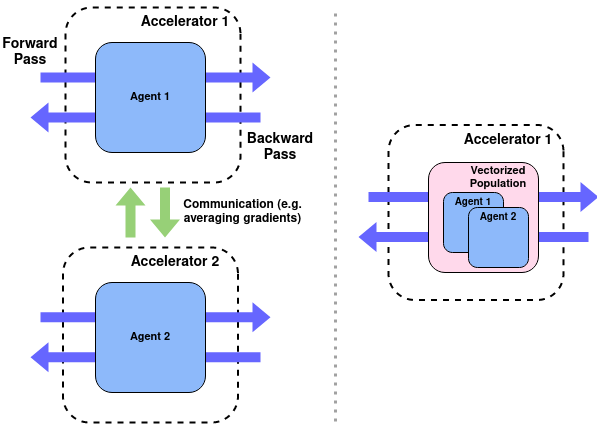}
    \caption{Schematic of two different parallelization implementations for population-based learning. The stereotypical implementation is depicted on the left where each agent is trained using a dedicated hardware accelerator. The right panel depicts another approach, which we revisit in this study, where computations are vectorized  over the population using a single accelerator, leveraging the parallelization capabilities of the hardware until saturation.}
    \label{fig:illustration-vectorization}
\end{figure}

In the past decade, Deep Learning and Deep Reinforcement Learning have led to breakthroughs in many fields such as computer vision, games, natural language processing, robotics, and bioinformatics. At a very high level, these methods all involve randomly initializing the parameters of neural networks and repeatedly applying the following steps until convergence: (1) computing a loss function over some training data, (2) differentiating it with respect to the parameters, and (3) updating them with a variant of stochastic gradient descent.

In Reinforcement Learning (RL), an agent learns to interact with an environment. At each time step, the agent receives an observation and takes, conditionally to this observation, an action. Once the action is carried out in the environment, the agent receives a reward as well as a new observation and the process repeats itself. The goal of the agent is to maximize the rewards it receives during these interactions. Deep policy search methods, a widely used class of Deep Reinforcement Learning methods, use a neural network to parametrize the agent's policy, i.e. the function that maps observations to actions. These algorithms alternate between two steps: data collection and policy update. Typically, the agent interacts with the environment for a fixed number of time steps after which its policy gets updated following the generic Deep Learning recipe by formulating a loss function and using the data that has been collected during the previous interactions. In many popular approaches, the data collected by the agent in the environment is not immediately discarded but instead stored inside a replay buffer, a data structure with a maximum size and a first-in first-out insertion rule, from which the agent can sample during subsequent update periods.

Recently, several works have showed that the performance of Deep Learning methods can be improved by training a population of neural networks at the same time instead of a single one. These techniques have yielded impressive results in multiple Deep Learning applications such as for model compression and hyperparameter tuning. Training a population was also demonstrated to be advantageous in Deep Reinforcement Learning applications where several policies can be exploited to enhance exploration, resulting in improved sample efficiency and/or asymptotic performance, or for skills discovery.

While training a population instead of a single individual is advantageous from a performance standpoint, this comes at a computational cost as the number of forward and backward passes through neural networks typically scales linearly with the population size. Consequently, naive sequential implementations tend to be prohibitively slow, especially for large populations. In the simplest settings, these computations can be performed independently in parallel across the population and thus be carried out just as fast as for a single neural network in theory. However, this entails either: (i) dedicating one accelerator to each agent, which quickly requires large computational infrastructures, possibly involving cross-communication between multiple machines, see Figure~\ref{fig:illustration-vectorization}, or (ii) spawning one process per agent and sharing the accelerator between processes, but this is typically inefficient as we show in Section \ref{sec-comparison-of-implementations}. The complexity of setting up large infrastructures or the slowness of sequential implementations may draw researchers and practitioners away from these methods, which remain accessible only to large labs or companies with significant resources.

In this work, we aim to demonstrate that, with a proper implementation, it is possible to train populations of neural networks on a single machine. Our contributions are the following. \textbf{1.} We compare, from the perspective of speed, cost, scalability, and ease of development, possible implementations of population-based learning approaches and show that the judicious use of vectorization, compilation, efficient data structures and data loading tools brings about significant benefits in applications where the number of trainable weights of the neural networks is relatively small, such as is typical in RL. \textbf{2.} We revisit prior population-based learning studies from the literature and show that, armed with these tools, we can achieve better performance-over-runtime profiles on a single machine than when training a single neural network. \textbf{3.} We publicly release our code\footnote{\href{https://github.com/instadeepai/Fastpbrl}{https://github.com/instadeepai/Fastpbrl}} in the hope that it will encourage practitioners to use population-based learning more frequently for their research and applications.


\section{Related Work}

\textbf{Population-based methods for Deep Learning applications} Population-based training methods have been applied successfully for model compression, a process commonly referred to as distillation, see \cite{hinton2015distilling}. In this line of work, the goal is to replace a single very large neural network with a family of smaller models that achieve similar performance through ensembling techniques. The training process typically involves running inferences of training data through the original large neural network and building loss functions to replicate the output of the some of the layers of the neural network (as opposed to simply the final output). In this setting, the smaller models are all initialized randomly with different weights and are updated independently. Population-based Deep Learning methods have also successfully been used for hyperparameter tuning to reach better performance. Notably, algorithms such as Population Based Training (\pbt) \cite{jaderberg2017population, vinyals2019grandmaster, jaderberg2019human}  train populations of models with different values for the hyperparameters and use a genetic algorithm to update the population regularly. 

\textbf{Population-Based Reinforcement Learning} One successful application of Population-Based Reinforcement Learning (PBRL) is to train multiple policies so as to generate diverse behaviors that can be used to warmstart another training or for fast adaptation in applications such as damage recovery \cite{eysenbach2018diversity, sharma2019dynamics, colas2020scaling}. Another booming application of PBRL is to use several policies as a way to explore more efficiently than by adding random noise to the actions when interacting with the environment in order to improve sample efficiency, convergence speed, training stability and even asymptotic performance in hard environments. This exploration can be performed in different spaces. Some methods, such as \dvd \cite{parker2020effective}, directly encourage the policies to visit different parts of the state space through a shared loss function. In contrast, some methods, such as \pssstddd \cite{jung2020population}, independently initialize and update the policies but penalize them if they deviate too much from the best one. Some other methods, such as \pgame \cite{Nilsson2021} or \qdpg \cite{pierrot2022diversity}, take a genetic approach to evolve the population with the objective of covering a space of behaviors (which characterize the interactions with the environment) with high-performing policies. Finally, other approaches, such as \cemrl \cite{pourchot2018cem}, directly explore in the parameters space by perturbing the weights randomly. \\ 

\textbf{Existing implementations and parallelization techniques}
Closest to our work are \cite{espeholt2018impala} and \cite{petrenko2020sample} as the focus in these studies is also to train Deep Reinforcement Learning agents on a single machine with a single hardware accelerator as efficiently as possible by fully leveraging the parallelization capabilities of the accelerator. However the focus in these prior works is different. The authors of \cite{espeholt2018impala} aim to train a single agent parametrized by a large and deep neural network on a challenging environment involving visual feedback. Compared to our work, vectorization is used over the batch size of the training data while we vectorize computations over multiple small neural networks. The authors of \cite{petrenko2020sample} focus on developing tools and strategies to speed up the data collection process (as opposed to the update steps in our work) for training a population of agents on visual tasks. The only strategy considered in this prior work to speed up the update steps across the population is to spawn multiple processes and share the accelerator, which we show is significantly less efficient than vectorizing in Section \ref{sec-comparison-of-implementations}. Many other prior works are concerned with improving the data collection process to be able to feed training data to the process in charge of updating the neural network parameters without delay, such as \cite{horgan2018distributed} and \cite{stooke2021decoupling}. We leverage these ideas in our implementations to guarantee that training data generation is never a bottleneck but this is not the focus of this work. Less recent prior work often relied on distributed asynchronous gradient descent to speed up the training process, as in \cite{mnih2016asynchronous}, but this is the setting we aim to avoid as this typically requires many hardware accelerators. However, we do compare this parallelization technique to the vectorization approach when the hardware accelerators used are regular central processing units (CPUs) as single machines can have up to 200 of them.

\section{Problem Statement}

We consider a finite-horizon Markov Decision Process (MDP), $\left( \mathcal{S}, \mathcal{A}, \mathcal{T}, \mathcal{R}, \gamma \right)$ where $\mathcal{S}$ is the state space, $\mathcal{A}$ is the action space, $\mathcal{T}: \mathcal{S} \times \mathcal{A} \rightarrow \mathcal{P}(\mathcal{S})$ is the transition dynamics (where $\mathcal{P}(\mathcal{S})$ is the set of probability distributions over $\mathcal{S}$), $\mathcal{R}: \mathcal{S} \times \mathcal{A} \rightarrow \reals$ is the reward function and $\gamma \in [0,1]$ is a discount factor. In this work, we consider policy search methods that seek for a policy $\pi_{\theta}: \mathcal{S} \rightarrow \mathcal{D}_{\mathcal{A}}$ parameterized by $\theta \in \Theta$ where $\Theta$ is referred to as parameters space and $\mathcal{D}_{\mathcal{A}}$ is the space of distributions over the action space.  In this paper, the policy is assumed to be a neural network. 

Standard (as opposed to population-based) Reinforcement Learning methods approaches aim to find one optimal policy $\pi_{\theta^*}$ that maximizes the discounted sum of rewards derived when interacting with the environment. In policy search methods, the policy is initialized with random parameters and updated progressively through stochastic gradient descent until convergence. In contrast, PBRL methods update a population of  $N$ policies $\{ \pi_{\theta_i} \}_{i=1,N}$ where each policy interacts with its own copy of the environment independently from the others (whereas in multi-agent RL, the agents would jointly interact with the environment). Population-based approaches typically vary along the exact way they update the population periodically, whether parameters are shared across the population, and whether the loss functions involve cross-agents terms. 

In this work we assume that the neural networks used as function approximators are relatively small, in the sense that: (1) the memory required to store their parameters is a fraction of the memory available on modern accelerators and (2) running a single batch of data of typical size through them uses only a fraction of the parallelization capabilities of modern accelerators. This is typically the case in many previous studies that achieve excellent performance in locomotion and robotic simulation environments \cite{fujimoto2018addressing, haarnoja2018soft} as we verify in Section \ref{sec-comparison-of-implementations}. Training large neural networks requires considerable additional effort, time and resources to stabilize the training process and obtain better performance \cite{ota2021training}. 

In addition, we assume that interacting with the environment for one timestep (i.e. running one policy network forward pass and stepping the environment simulation one time) is fast compared to carrying out one update step for the agent. This is typically the case in locomotion and robotic simulation environments considered in the literature as we discuss in Section \ref{sec:data-collection} of the Appendix. 


\section{Comparison of Implementations}
\label{sec-comparison-of-implementations}

In this section, we assume that a single hardware accelerator is available and we compare multiple implementations of the update step of a population of agents assuming that batches of training data that have already been loaded in the accelerator memory are available without delay upon request. In particular, we show that vectorizing computations that occur as part of agents' updates over the population can be up to two orders of magnitude faster than training the agents sequentially. In this work, we focus on optimizing the update step time rather than the data collection process as the use of hardware accelerators to speed up the computations is not supported for many commonly-used simulation environments, such as Gym \cite{openaigym} with a MuJoCo backend \cite{todorov2012MuJoCo}, and we want the discussion to be as generic as possible. In Section \ref{sec:data-collection} of the Appendix, we discuss simple strategies for efficient data collection of environment interactions to guarantee that this does not become a bottleneck when leveraging vectorization for training the agents, even with limited computational resources. These strategies might fall short for simulation environments with high-dimensional observation spaces, such as for environments with visual inputs, or long step times, in which case more involved strategies, such as those described in \cite{petrenko2020sample}, or simulators with built-in support for hardware accelerators, such as Brax \cite{brax2021github} or Isaac Gym \cite{makoviychuk2021isaac}, must be used to guarantee that gathering training data does not become a bottleneck.

We first focus on the setting considered in \cite{jaderberg2017population} where $N$ replicas of the same agent are trained independently but with different initialization of the parameters and different hyperparameters. This is the simplest possible PBRL setting which will enable us to describe possible implementations as simply as possible and illustrate the benefits of (automated) vectorization. Note that this framework also includes as a special case running many seeds of the same agent for benchmarking purposes, which, in light of recent studies \cite{agarwal2021deep}, should be done more systematically with a large number of seeds to avoid drawing flawed conclusions. In Section \ref{sec:training-shared-params}, we extend the study to a more common - albeit slightly more complicated - setting where either some neural network parameters are shared across the population of agents or some terms in the loss functions involve all policies. This is where automated vectorization frameworks really shine as parallelizing the training of the agents is no longer straightforward to implement (gradients need to flow between accelerators).


\subsection{Training Independent Replicas of the Same Agent}
\label{sec:training-same-agent} 

We use three standard off-policy RL algorithms in our experiments: Soft Actor Critic \cite{haarnoja2018soft} (\sac) and Twin Delayed Deep Deterministic Policy Gradient (\tddd) \cite{fujimoto2018addressing}, with the standard hyperparametrization used in previous studies \cite{haarnoja2018soft} for MuJoCo Gym locomotion environments, as well as Deep Q-Learning (\dqn), with the original hyperparametrization used for the Atari 2600 games from the Arcade Learning Environment \cite{mnih2013playing}. Specifically, we use fully-connected neural networks (resp. convolutional neural networks followed by fully-connected layers) to parametrize the critics and policies of \sac and \tddd (resp. the critic of \dqn). For the empirical study of this section on \sac and \tddd, we generate training data corresponding to training agents on the MuJoCo Gym HalfCheetah-v2 environment but similar results can be derived for robotic and locomotion simulation environments with higher dimensional observation and action spaces such as Humanoid-v2. For \dqn, we use the same pipeline used in \cite{mnih2013playing} to preprocess and stack images from the Atari 2600 games.

\paragraph{Description of possible implementations}
We compare the following implementations for carrying out one update step for a population of $N$ replicas of the same agent (same architecture but with different weights).

\textbf{Torch (Sequential)} A sequential implementation, referred to as \emph{Torch (Sequential)}, where we carry out one update step for each agent sequentially with  \torch, a dynamic computational graph deep learning library \cite{NEURIPS2019_9015}. This is a widely used approach in the PBRL literature, see for example \cite{pourchot2018cem} and \cite{parker2020effective}, as it is flexible and easy to write.

\textbf{Jax (Sequential)} A sequential implementation, referred to as \emph{Jax (Sequential)}, where we carry out one update step for each agent sequentially with \jax \cite{frostig2018compiling, jax2018github}, a deep learning library leveraging Just-In-Time (\jit) compilation for accelerators. Compared to the dynamic computational graph approach used by default in \torch, \jax builds an accelerator-specific compiled version of the static graph of computations the first time the graph is run. This is more efficient when the same graph of computation is used many times in a row with different inputs, as is typical when training RL agents. \torch also provides the option to \jit-compile computational graphs with the TorchScript module but support for \jit compiling backward passes through neural networks remains limited at the time of writing.

\textbf{Torch (Vectorized)} A vectorized implementation, referred to as \emph{Torch (Vectorized)}, where we use \torch and the intermediate layers of the neural networks of the agents are expanded and concatenated along the first dimension beforehand to build a single neural network model with $N$ times the number of parameters. We leverage vectorized operations implemented in the framework, such as \emph{matmul} for batch matrix-matrix products and \emph{Conv2d} with the \emph{groups} option for batch convolutions. This is illustrated with a code snippet in Section \ref{sec-example-vectorization-code} of the Appendix.

\textbf{Jax (Vectorized)} A vectorized implementation, referred to as \emph{Jax (Vectorized)}, where we leverage the vectorization primitive \emph{vmap} from \jax by wrapping it around the update step function defined for a single agent and \jit compiling the result. For convolutional neural networks, we use the \emph{feature\_group\_count} option of the \emph{Conv2d} module from the haiku library \cite{haiku2020github} to vectorize computations in a similar fashion as done with \torch as it turns out to be more efficient than \emph{vmap}.

\textbf{Torch (Parallel)} A parallel implementation, referred to as \emph{Torch (Parallel)}, where we use \torch and we spawn one process per agent in the population. All processes use the same accelerator concurrently and independently.

\textbf{Jax (Parallel)} A parallel implementation, referred to as \emph{Jax (Parallel)}, where we use \jax and we spawn one process per agent in the population. All processes use the same accelerator concurrently and independently.

We use the state-of-the-art implementations of \sac, \tddd, and \dqn from the \acme library \cite{hoffman2020acme} for the approaches relying on the \jax backend and the ones from Stable-Baselines3 \cite{stable-baselines3} for the approaches relying on the \torch backend. In all cases we use the latest available version of the library available, the corresponding version of \cuda, and the latest available driver version for each hardware accelerator.

For all of these implementations, we also consider another variant where we carry out 50 (resp. 10) update steps in a row for \tddd and \sac (resp. \dqn) without copying the values of the neural network parameters to the host memory between update steps. This is a commonly used asynchronous technique developed to speed up the computations with minimal impact on performance where multiple update steps are carried out before updating the actors with new values for the neural network parameters (but with the ratio of number of update steps over number of environment interactions still constrained around a target value). This is more efficient for two reasons: (1) copying data from the hardware accelerator to the host memory takes time so many deep learning software frameworks (such as \torch and \jax) send execution traces to the hardware accelerator(s) but do not wait for their completions until the values of some of the parameters involved are accessed and (2) this reduces the communication overhead between the actors and the process(es) in charge of carrying out the update steps.

\begin{figure*}[ht!]
    \centering
    \includegraphics[width=0.98\textwidth]{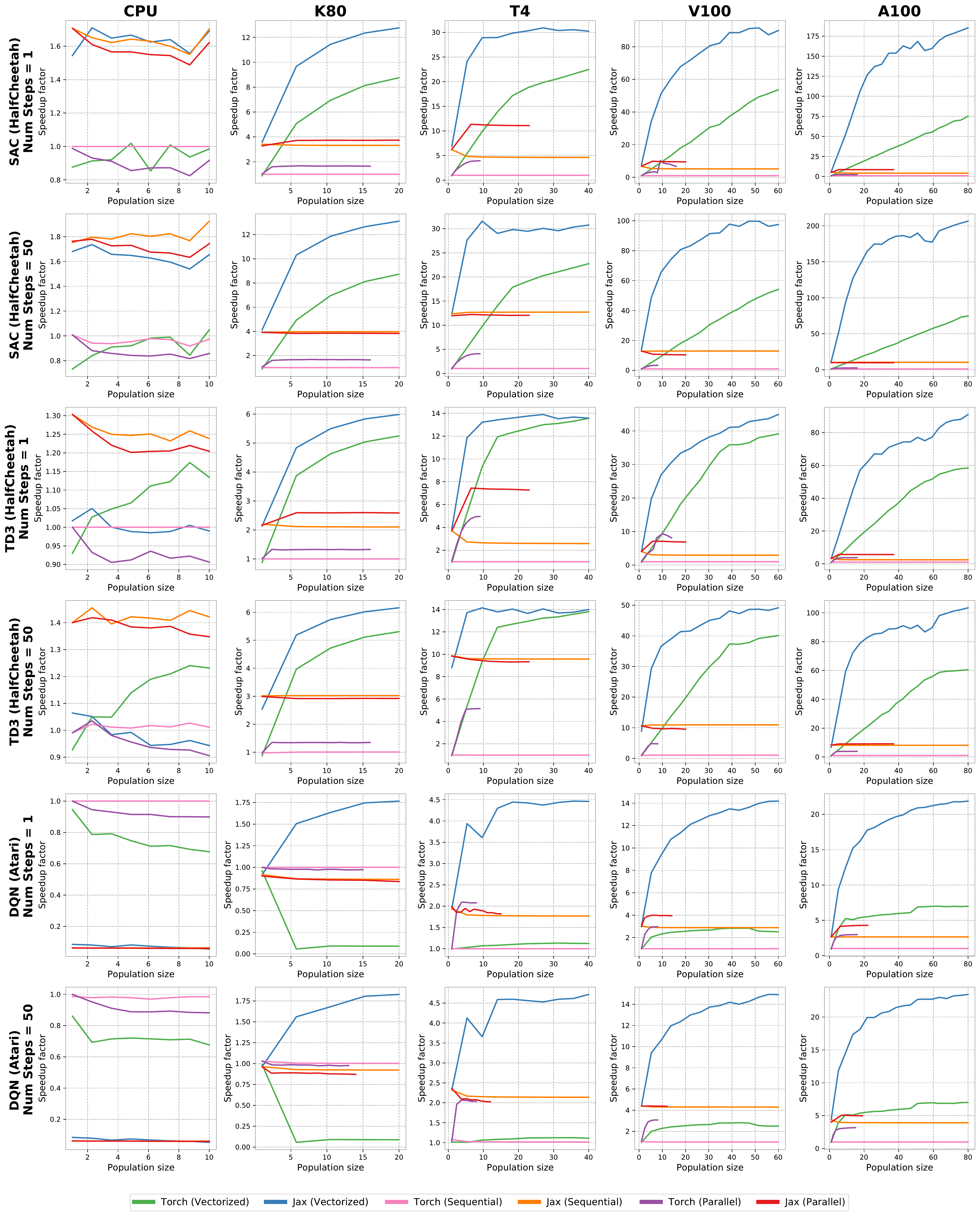}
    \caption{Parameters update speed-up study for various implementations and accelerators for the \sac and \tddd (resp. \dqn) agents on the HalfCheetah-v2 (resp. Atari 2600) environment. All of the implementations are compared to the \emph{Torch (Sequential)} implementation (with num\_steps = 1)  and speed-up factors are reported w.r.t. this implementation. \emph{CPU} refers to a single thread core of an Intel Xeon 2.80Ghz processor.}
    \label{fig:results_runtime_speedup}
\end{figure*}

\textbf{Numerical experiments}
In Figure \ref{fig:results_runtime_speedup}, we benchmark the time it takes to carry out one update step as a function of the population size $N$ given $N$ data batches that have been previously loaded in the memory of the hardware accelerator (or in \ram memory if we are using a \cpu). The time it takes to load the training data on the accelerator is not taken into account as it can be done in an asynchronous fashion with little to no overhead as detailed in Section \ref{sec:data-collection} of the Appendix. If the approach causes an out-of-memory error for a given population size (which happens only for the parallel approaches as discussed below), the corresponding update step time is not reported on Figure \ref{fig:results_runtime_speedup}.

We observe that the capabilities of modern hardware accelerators are heavily under-utilized when training \sac, \tddd, or \dqn agents sequentially with the neural network parametrizations typically considered in previous studies. For population sizes larger than 15, vectorizing computations with \jax leads to a speed-up factor of at least 4 compared to using the sequential \torch implementation for all but one of the hardware accelerators considered. This factor increases with the parallelization capabilities of the hardware accelerator, and almost reaches double-digit range for population sizes of 80 for the latest available hardware. Additionally, \emph{Jax (Vectorized)} is always faster, often by a large margin, than \emph{Jax/Torch (Parallel)} for all accelerators. 

The results also show that a 2 to 14 speedup factor can be readily obtained by compiling the static graph of computations for a moderate initial compilation time, see Table \ref{table-compilation-time} in the Appendix. In addition, as the population size increases, the advantage provided by compiling the graph of computations diminishes because the time it takes to dynamically recalculate the graph of computations and push it to the accelerator becomes negligible compared to the time it takes to carry out the computations (compare \emph{Torch (Vectorized)} to \emph{Jax (Vectorized)} on Figure \ref{fig:results_runtime_speedup}). Finally, vectorization brings no speedup when using one \cpu core only. Surprisingly, the \jax-based implementations are much slower than the \torch-based ones on \cpu for \dqn on Atari, perhaps because the underlying kernel is not optimized for this combination of hardware and neural network architecture.



\begin{figure*}[ht!]
    \centering
    \includegraphics[width=0.95\textwidth]{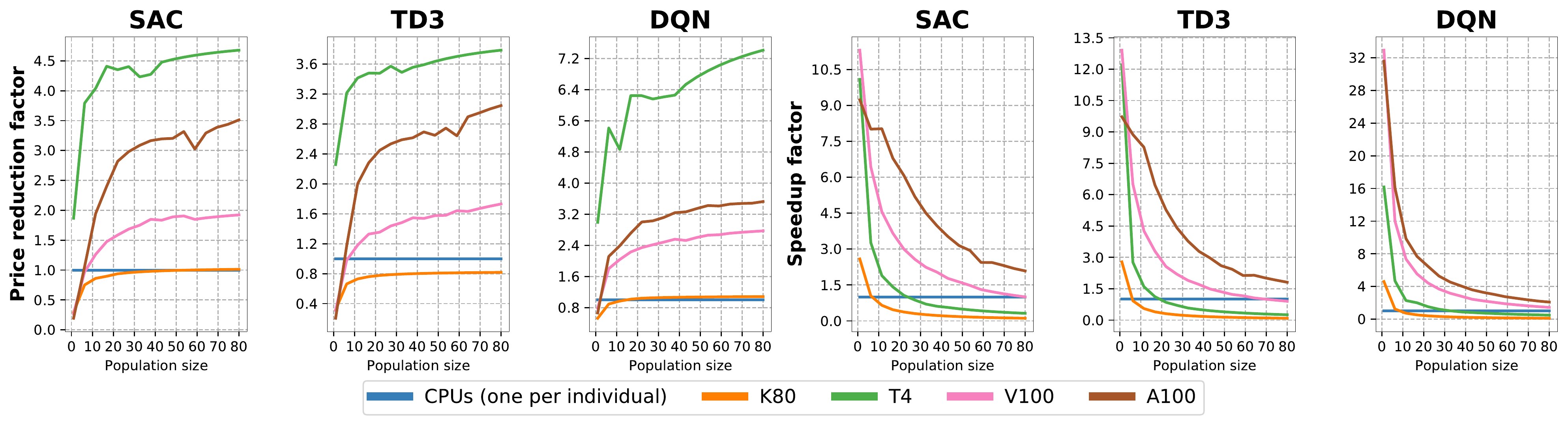}
    \caption{Comparative cost and runtime analysis for various hardware accelerators as a function of the population size. All of the accelerators are compared to using CPUs (one per agent in the population) and metrics are reported w.r.t. this choice of hardware. Costs per second for the accelerators are derived by averaging the posted prices from three main cloud computing platforms, see Table \ref{table-posted-price}.}
    \label{fig:cost_analysis}
\end{figure*}

While Figure \ref{fig:results_runtime_speedup} shows that the combination of vectorization and compilation can yield significant speedups for moderate to large population sizes when using hardware accelerators, one might still wonder if training independent agents in parallel on \cpus could be more cost-effective and/or faster as the runtime does not vary with population size if we allocate one core per agent. Note that this strategy is straightforward to implement when training independent agents but less so when some parameters are shared between agents as discussed in Section \ref{sec:training-shared-params}. Based on the observations above, we restrict the comparative study to the \emph{Jax (Vectorized)} implementation when using hardware accelerators and the \emph{Jax (Sequential)} (resp. \emph{Torch (Sequential)}) implementation for \sac and \tddd (resp. \dqn) when using \cpu cores (one per agent in the population), \jit-compiling 50 (resp. 10) update steps in both cases. Surprisingly, it turns out that, for any population size between 1 and 80, there is always at least one hardware accelerator considered in this study that performs better both in terms of speed and cost than using one \cpu core per agent, as shown in Figure  \ref{fig:cost_analysis}. On the other hand, as expected, no hardware accelerator is both faster and cheaper than any other for all population sizes so practitioners can pick and choose based on their own cost/runtime trade-off. 

\textbf{Memory considerations}
Vectorizing update steps across the population increases the accelerator's effective memory usage compared to sequentially updating the agents as intermediate computations for all agents need to be stored concurrently in the accelerator's memory. This can impose a practical limit on the maximum population size for vectorized implementations, especially for large neural networks, although this limit is beyond the point where the parallelization capabilities of the hardware are fully utilized in our experimental study, as can be seen by the fact that the speedup factors level off before we reach the maximum population sizes considered in this study in Figure \ref{fig:results_runtime_speedup}. Moreover, we stress that the effective memory usage of the vectorized approach is sublinear w.r.t. the population size because (1) allocating memory in a single chunk for the entire population minimizes memory fragmentation and (2) CUDA kernels are loaded only once. This is contrary to the parallel approach where memory usage is in fact proportional to the population size, which severely limits the maximum population size that this approach can accommodate in our experiments, as can be seen on Figure \ref{fig:results_runtime_speedup}.

\subsection{Training Agents Sharing Parameters}
\label{sec:training-shared-params}

\begin{figure}[t]
    \centering
    \includegraphics[width=0.4\textwidth]{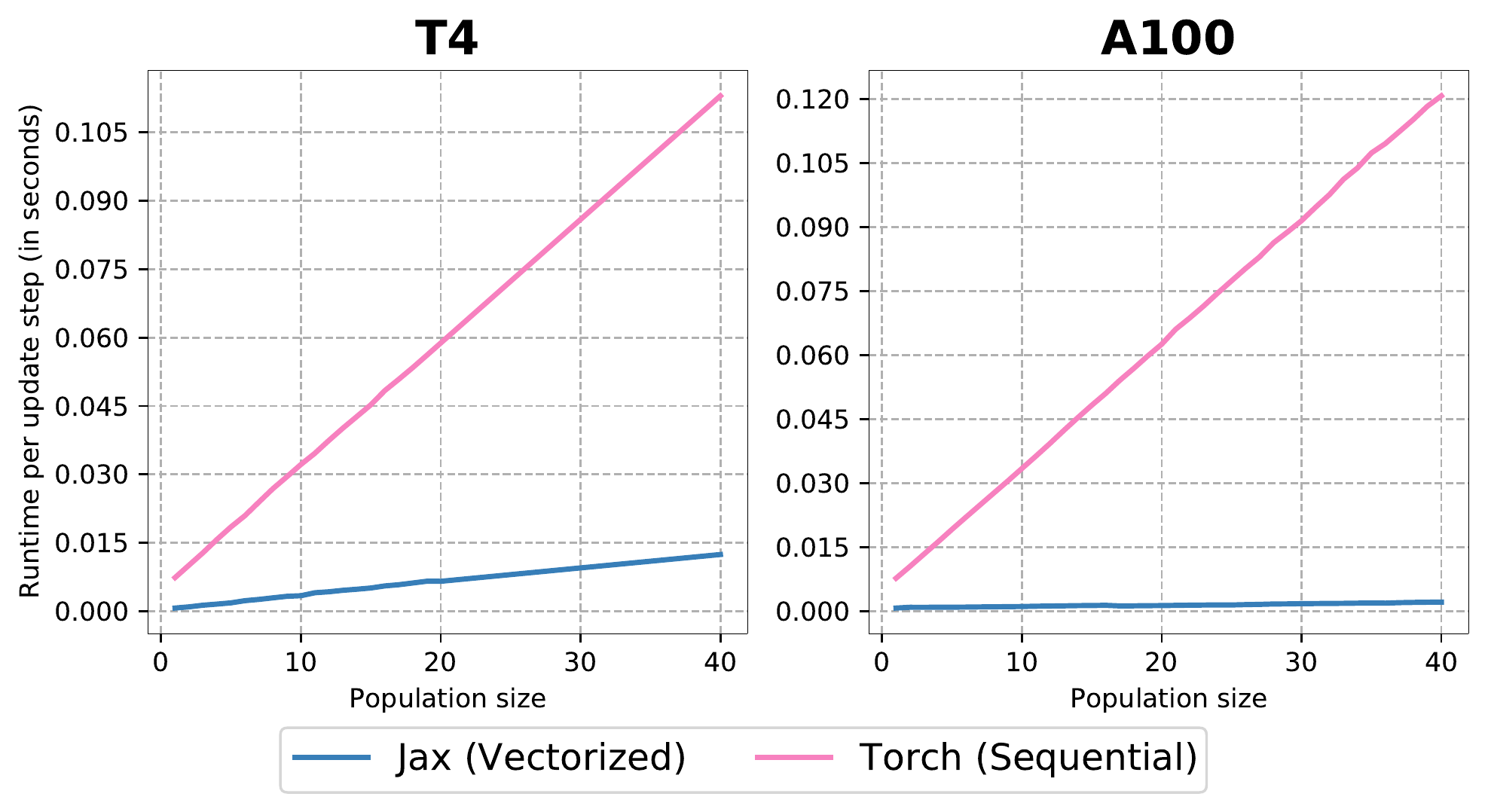}
    \caption{Runtime comparison of one update step for a population of TD3 agents sharing critic parameters as a function of the population size for various hardware accelerators and implementations.}
    \label{fig:shared-critic-runtime}
\end{figure}

Many PBRL frameworks, such as \cemrl \cite{pourchot2018cem}, \dvd \cite{parker2020effective} and \qdpg\cite{pierrot2022diversity}, share parameters between agents of the population to improve the performance (in terms of cumulative sum of rewards per episode) achieved for a given number of interactions with the environment simulation. This makes strategies that scale the number of agents with the number of accelerators harder to implement in practice as this requires the efficient use of an inter-process communication protocol at every update step to aggregate gradients with respect to the shared parameters across the entire population. The use of automated vectorization primitives in specialized libraries, such as in \jax, makes this trivial to implement but this may require to vectorize the pseudo-code of the update step function itself as a pre-requisite. As a practical example, we consider the strategy used in \cemrl, \dvd, and \qdpg for \tddd where the critic parameters are shared between all agents in the population while each agent has its own set of policy parameters. The \tddd critic and policy loss functions considered in these works are the same as for the single-agent version. Additionally, given the shared critic parameters, the policy parameters undergo the same update as for the single-agent version. However, updates of the critic parameters are intertwined between the sequential updates of the policy parameters as illustrated in Algorithm 1 of \cemrl. The sequential nature of these updates prevents the vectorization of computations over the population so we consider a second-order modification where the critic overall undergoes the same number of updates but each batch of training data goes through all of the policy networks in parallel and the critic loss is averaged over the population. This approach can easily be implemented in \jax and leads to significant speedups compared to the sequential approach put forth in these papers, as shown in Figure \ref{fig:shared-critic-runtime}. We show that this change does not impact asymptotic performance or sample efficiency by replicating the experiments done in previous studies in Sections \ref{sec-cem-rl} and \ref{sec-dvd}.

\section{Case Studies}
\label{sec-case-studies}
In this section, we revisit three previous PBRL studies and showcase the benefits brought about by the combined use of vectorization and JIT compilation when training on a single machine. For all studies, the performance is reported w.r.t. the overall wall time, which not only includes the total time spent running update steps for the population but also the total time spent waiting for agents to interact with the environment, although the latter accounts for a small fraction of the overall wall time as we efficiently parallelize data collection across \cpu cores, as described in Section \ref{sec:data-collection}.

\subsection{Hyperparameter Tuning}
\label{sec-hyperparameter-tuning}

We first revisit a notoriously computationally expensive problem: tuning the hyperparameters of RL agents. The authors of \cite{jaderberg2017population} show that tuning the hyperparameters to the environment at hand can yield substantial improvements (in terms of asymptotic performance) compared to using the same set of hyperparameters that have been shown to perform well for a variety of environments. At a high level, the hyperparameter tuning process described in \cite{jaderberg2017population}, \pbt, runs in iterations where agents are trained independently for a significant number of update steps. At the end of each iteration, the worst performing agents are discarded and new ones are generated to replace them by copying the parameters of randomly sampled agents among the best performing ones. For these agents, a new set of hyperparameters is generated, either by sampling from a prior distribution or by perturbing the hyperparameter values of the parent agent. While \cite{jaderberg2017population} is a large-scale study leveraging computing clusters aimed at attaining state-of-the-art performance for complex environments involving visual feedback, we aim to show that it is possible to run similar experiments with the same population sizes on simpler environments on a single machine. Specifically, we train populations of 80 \tddd (resp. \sac) agents on 4 T4 hardware accelerators, each using the \emph{Jax (Vectorized)} implementation for training 20 agents in parallel. We could have used a single A100 hardware accelerator for the same total runtime but using 4 T4 accelerators in parallel is comparatively cheaper on cloud computing platforms, see Figure \ref{fig:cost_analysis}. Exact details of the experiments are reported in Section \ref{sec-experimental-details-pbt} of the Appendix.

\begin{figure*}[ht!]
    \centering
    \includegraphics[width=0.98\textwidth]{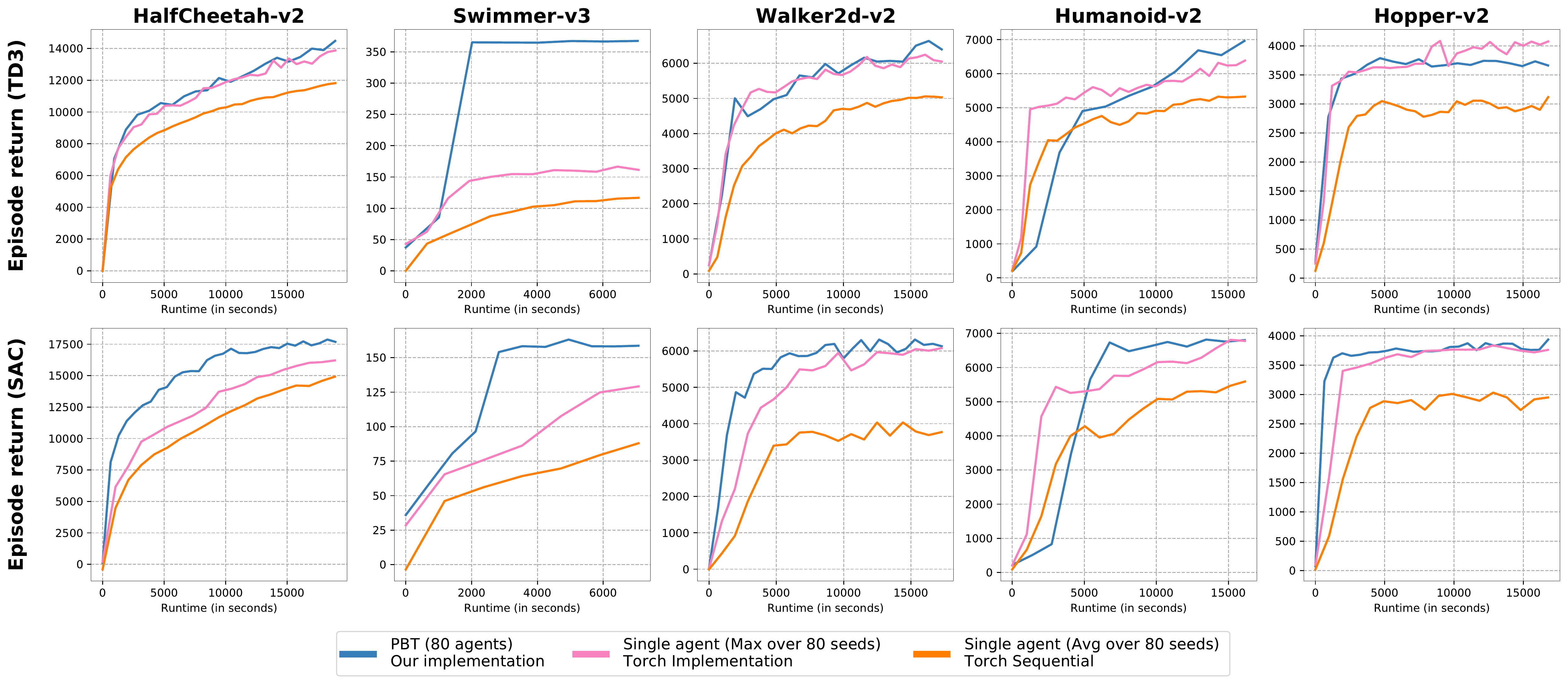}
    \caption{Evaluation of our \pbt implementation. Performance (in terms of mean episode returns) achieved as a function of total time elapsed since the beginning of the training run for various implementations and Gym locomotion environments. All experiments are run on a single machine with 4 T4 accelerators and 40 \cpu cores.}
    \label{fig:results_pbt}
\end{figure*}

We evaluate the performance of our implementation by looking at the episode return achieved by the best agent in the population as a function of the total time elapsed since the beginning of the training run. In Figure \ref{fig:results_pbt}, we compare this to the same metric (averaged over 80 seeds and by taking the maximum over 80 seeds) measured for a single agent implemented using a state-of-the-art \torch implementation. The single agent uses a single T4 accelerator as splitting batches of training data over multiple accelerators did not yield speed-up improvements (dynamic computational graph compilation costs seem to dominate for small neural networks and batch sizes). As seen in Figure \ref{fig:results_pbt}, for any given time elapsed since the beginning of the training, we achieve often better, sometimes on par, and rarely slightly worse, performance than the best agent out of 80 seeds. Interestingly, the optimization of the hyperparameters does not always lead to an improvement in asymptotic performance, as the initial hyperparameters used come from prior studies that have refined them over time, but sometimes yields faster learning that cannot be explained alone by the variability in performance seen over 80 different seeds of the same agent, see Figure \ref{fig:pbt_steps} of the Appendix.

\begin{figure*}[ht!]
    \centering
    \includegraphics[width=0.784\textwidth]{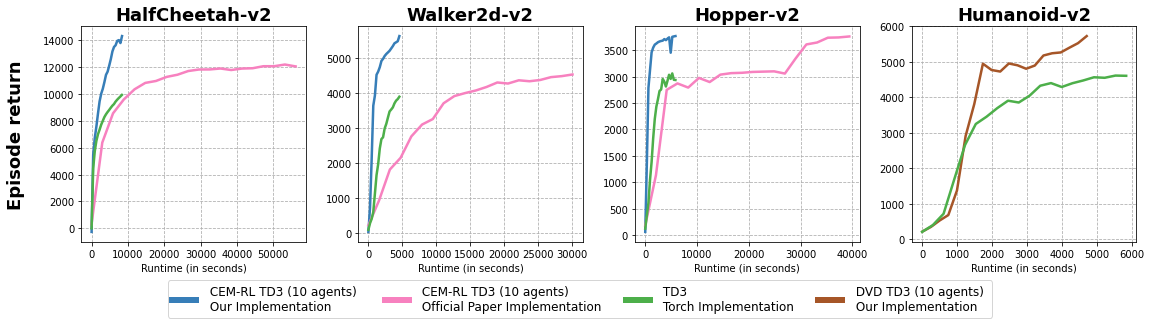}
    \caption{Evaluation of our \cemrl and \dvd implementations. Performance (in terms of mean episode returns) achieved as a function of total time elapsed since the beginning of the training run for various implementations and Gym locomotion environments. All experiments are run on a single machine with a T4 accelerator and 16 \cpu cores.}
    \label{fig:cem_dvd_runtime}
\end{figure*}

\subsection{CEM-RL}
\label{sec-cem-rl}

We now revisit another PBRL study, namely \cite{pourchot2018cem}, where parameters are shared across the agents and parallelization of the update step over the population is thus no longer trivial to implement. The authors of this work propose to combine the Cross-Entropy-Method (\cem), an Evolutionary algorithm, with off-policy RL algorithms, among which \tddd. The approach maintains independent distributions for all policy parameters and agents are generated by sampling from these distributions. Half of the population thus generated undergoes the standard \tddd update function for a number of steps, using a single set of critic parameters shared across the population but different sets of policy parameters for the agents, while the other half is left untouched and only evaluated. The authors use the \cem algorithm to update the means and variances of the distributions of the parameters using only the top half of the population ranked in terms of episode returns.

We replicate the study carried out in  \cite{pourchot2018cem} on the same environments and with the same population size of 10 as in the original work but we use the \emph{Jax (Vectorized)} implementation whereas the original authors use the \emph{Torch (Sequential)} one. Both implementations run on the same hardware (1 T4 accelerator with 16 \cpu cores) but we note that our implementation also parallelizes the interactions with the environment, although the total environment step time accounts only for a fraction of the total runtime of the original implementation so this strategy by itself yields little speedup in this case. As detailed in Section~\ref{sec:training-shared-params}, we make a small modification to the training process originally put forth in \cemrl to enable vectorization of the update step over the population. The experimental results shown in Figure~\ref{fig:cemrl_dvd_steps} of the Appendix show that this is not detrimental to the sample-efficiency of the approach.

We evaluate the performance of our implementation by looking at the episode return achieved by the policy network corresponding to the average of the policy network distribution maintained by \cem (as in the original study) as a function of the total time elapsed since the beginning of the training run and do the same for the original implementation released by the authors. Just like for Section \ref{sec-hyperparameter-tuning}, we also compare our implementation on the basis of the same metric (averaged over 80 seeds) measured for a single agent implemented using a state-of-the-art \torch implementation. As seen in Figure~\ref{fig:cem_dvd_runtime}, the combination of vectorization and compilation yields major speedups compared to a sequential implementation,  which confirms the benefit of our approach even for small population sizes.

\subsection{DvD}
\label{sec-dvd}

For this last case study, we consider \dvd \cite{parker2020effective}, which has the same goal of improving the sample efficiency of a \tddd agent as \cemrl. Compared to the latter, this is achieved by introducing an additive term in the loss function that explicitly favors diversity in the actions chosen by the policy networks. This additive term is expressed as the volume of an abstract embedding space of the policy networks and it jointly depends on the policy parameters of all the agents in the population, further complicating the process of parallelizing update steps across agents. This extension is trivial to implement with \jax building upon the \cemrl one as the policy parameters of all the agents are jointly stored in the same matrices. We replicate the original study on the Humanoid-v2 environment by training a population of 5 \tddd agents on one T4 accelerator.

Similarly as for the other two case studies, we evaluate the performance of our implementation by looking at the best of the agent in the population as a function of the total time elapsed since the beginning of the training run and we compare this to the same metric (averaged over 80 seeds) measured for a single agent implemented using a state-of-the-art \torch implementation in Figure \ref{fig:cem_dvd_runtime}. The approach is not compared with the original implementation released by the authors as it is prohibitively slow.

\section{Conclusion}
We have showed that, for small neural networks typically used in RL, the judicious use of vectorization combined with modern hardware-specific acceleration frameworks allows to run population-based RL algorithms on a single machine with a runtime comparable to running a single agent naively while reaching comparable or better asymptotic performance. The practical tips, analysis as well as code we provide can be used to extend our results and implementations to many other population-based methods beyond Reinforcement Learning. We hope that this work will benefit the community by allowing them to run population-based methods on affordable infrastructure in reasonable runtimes.

\newpage
\newpage

\bibliography{main}

\begin{thebibliography}{33}
\providecommand{\natexlab}[1]{#1}
\providecommand{\url}[1]{\texttt{#1}}
\expandafter\ifx\csname urlstyle\endcsname\relax
  \providecommand{\doi}[1]{doi: #1}\else
  \providecommand{\doi}{doi: \begingroup \urlstyle{rm}\Url}\fi

\bibitem[Agarwal et~al.(2021)Agarwal, Schwarzer, Castro, Courville, and
  Bellemare]{agarwal2021deep}
Agarwal, R., Schwarzer, M., Castro, P.~S., Courville, A.~C., and Bellemare, M.
\newblock Deep reinforcement learning at the edge of the statistical precipice.
\newblock \emph{Advances in Neural Information Processing Systems}, 34, 2021.

\bibitem[Bradbury et~al.(2018)Bradbury, Frostig, Hawkins, Johnson, Leary,
  Maclaurin, Necula, Paszke, Vander{P}las, Wanderman-{M}ilne, and
  Zhang]{jax2018github}
Bradbury, J., Frostig, R., Hawkins, P., Johnson, M.~J., Leary, C., Maclaurin,
  D., Necula, G., Paszke, A., Vander{P}las, J., Wanderman-{M}ilne, S., and
  Zhang, Q.
\newblock {JAX}: composable transformations of {P}ython+{N}um{P}y programs,
  2018.
\newblock URL \url{http://github.com/google/jax}.

\bibitem[Brockman et~al.(2016)Brockman, Cheung, Pettersson, Schneider,
  Schulman, Tang, and Zaremba]{openaigym}
Brockman, G., Cheung, V., Pettersson, L., Schneider, J., Schulman, J., Tang,
  J., and Zaremba, W.
\newblock Openai gym, 2016.

\bibitem[Colas et~al.(2020)Colas, Madhavan, Huizinga, and
  Clune]{colas2020scaling}
Colas, C., Madhavan, V., Huizinga, J., and Clune, J.
\newblock Scaling map-elites to deep neuroevolution.
\newblock In \emph{Proceedings of the 2020 Genetic and Evolutionary Computation
  Conference}, pp.\  67--75, 2020.

\bibitem[Espeholt et~al.(2018)Espeholt, Soyer, Munos, Simonyan, Mnih, Ward,
  Doron, Firoiu, Harley, Dunning, et~al.]{espeholt2018impala}
Espeholt, L., Soyer, H., Munos, R., Simonyan, K., Mnih, V., Ward, T., Doron,
  Y., Firoiu, V., Harley, T., Dunning, I., et~al.
\newblock Impala: Scalable distributed deep-rl with importance weighted
  actor-learner architectures.
\newblock In \emph{International Conference on Machine Learning}, pp.\
  1407--1416. PMLR, 2018.

\bibitem[Eysenbach et~al.(2019)Eysenbach, Gupta, Ibarz, and
  Levine]{eysenbach2018diversity}
Eysenbach, B., Gupta, A., Ibarz, J., and Levine, S.
\newblock Diversity is all you need: Learning skills without a reward function.
\newblock In \emph{International Conference on Learning Representations}, 2019.

\bibitem[Freeman et~al.(2021)Freeman, Frey, Raichuk, Girgin, Mordatch, and
  Bachem]{brax2021github}
Freeman, C.~D., Frey, E., Raichuk, A., Girgin, S., Mordatch, I., and Bachem, O.
\newblock Brax - a differentiable physics engine for large scale rigid body
  simulation, 2021.
\newblock URL \url{http://github.com/google/brax}.

\bibitem[Frostig et~al.(2018)Frostig, Johnson, and Leary]{frostig2018compiling}
Frostig, R., Johnson, M.~J., and Leary, C.
\newblock Compiling machine learning programs via high-level tracing.
\newblock \emph{Systems for Machine Learning}, 2018.

\bibitem[Fujimoto et~al.(2018)Fujimoto, Hoof, and
  Meger]{fujimoto2018addressing}
Fujimoto, S., Hoof, H., and Meger, D.
\newblock Addressing function approximation error in actor-critic methods.
\newblock In \emph{International conference on machine learning}, pp.\
  1587--1596. PMLR, 2018.

\bibitem[Haarnoja et~al.(2018)Haarnoja, Zhou, Hartikainen, Tucker, Ha, Tan,
  Kumar, Zhu, Gupta, Abbeel, et~al.]{haarnoja2018soft}
Haarnoja, T., Zhou, A., Hartikainen, K., Tucker, G., Ha, S., Tan, J., Kumar,
  V., Zhu, H., Gupta, A., Abbeel, P., et~al.
\newblock Soft actor-critic algorithms and applications.
\newblock \emph{arXiv preprint arXiv:1812.05905}, 2018.

\bibitem[Hennigan et~al.(2020)Hennigan, Cai, Norman, and
  Babuschkin]{haiku2020github}
Hennigan, T., Cai, T., Norman, T., and Babuschkin, I.
\newblock {H}aiku: {S}onnet for {JAX}, 2020.
\newblock URL \url{http://github.com/deepmind/dm-haiku}.

\bibitem[Hinton et~al.(2015)Hinton, Vinyals, and Dean]{hinton2015distilling}
Hinton, G., Vinyals, O., and Dean, J.
\newblock Distilling the knowledge in a neural network.
\newblock \emph{arXiv preprint arXiv:1503.02531}, 2015.

\bibitem[Hoffman et~al.(2020)Hoffman, Shahriari, Aslanides, Barth-Maron,
  Behbahani, Norman, Abdolmaleki, Cassirer, Yang, Baumli, Henderson, Novikov,
  Colmenarejo, Cabi, Gulcehre, Paine, Cowie, Wang, Piot, and
  de~Freitas]{hoffman2020acme}
Hoffman, M., Shahriari, B., Aslanides, J., Barth-Maron, G., Behbahani, F.,
  Norman, T., Abdolmaleki, A., Cassirer, A., Yang, F., Baumli, K., Henderson,
  S., Novikov, A., Colmenarejo, S.~G., Cabi, S., Gulcehre, C., Paine, T.~L.,
  Cowie, A., Wang, Z., Piot, B., and de~Freitas, N.
\newblock Acme: A research framework for distributed reinforcement learning.
\newblock \emph{arXiv preprint arXiv:2006.00979}, 2020.
\newblock URL \url{https://arxiv.org/abs/2006.00979}.

\bibitem[Horgan et~al.(2018)Horgan, Quan, Budden, Barth-Maron, Hessel, van
  Hasselt, and Silver]{horgan2018distributed}
Horgan, D., Quan, J., Budden, D., Barth-Maron, G., Hessel, M., van Hasselt, H.,
  and Silver, D.
\newblock Distributed prioritized experience replay.
\newblock In \emph{International Conference on Learning Representations}, 2018.

\bibitem[Jaderberg et~al.(2017)Jaderberg, Dalibard, Osindero, Czarnecki,
  Donahue, Razavi, Vinyals, Green, Dunning, Simonyan,
  et~al.]{jaderberg2017population}
Jaderberg, M., Dalibard, V., Osindero, S., Czarnecki, W.~M., Donahue, J.,
  Razavi, A., Vinyals, O., Green, T., Dunning, I., Simonyan, K., et~al.
\newblock Population-based training of neural networks.
\newblock \emph{arXiv preprint arXiv:1711.09846}, 2017.

\bibitem[Jaderberg et~al.(2019)Jaderberg, Czarnecki, Dunning, Marris, Lever,
  Castaneda, Beattie, Rabinowitz, Morcos, Ruderman, et~al.]{jaderberg2019human}
Jaderberg, M., Czarnecki, W.~M., Dunning, I., Marris, L., Lever, G., Castaneda,
  A.~G., Beattie, C., Rabinowitz, N.~C., Morcos, A.~S., Ruderman, A., et~al.
\newblock Human-level performance in 3d multiplayer games with population-based
  reinforcement learning.
\newblock \emph{Science}, 364\penalty0 (6443):\penalty0 859--865, 2019.

\bibitem[Jung et~al.(2020)Jung, Park, and Sung]{jung2020population}
Jung, W., Park, G., and Sung, Y.
\newblock Population-guided parallel policy search for reinforcement learning.
\newblock In \emph{International Conference on Learning Representations}, 2020.

\bibitem[Makoviychuk et~al.(2021)Makoviychuk, Wawrzyniak, Guo, Lu, Storey,
  Macklin, Hoeller, Rudin, Allshire, Handa, and State]{makoviychuk2021isaac}
Makoviychuk, V., Wawrzyniak, L., Guo, Y., Lu, M., Storey, K., Macklin, M.,
  Hoeller, D., Rudin, N., Allshire, A., Handa, A., and State, G.
\newblock Isaac gym: High performance gpu-based physics simulation for robot
  learning, 2021.

\bibitem[Mnih et~al.(2013)Mnih, Kavukcuoglu, Silver, Graves, Antonoglou,
  Wierstra, and Riedmiller]{mnih2013playing}
Mnih, V., Kavukcuoglu, K., Silver, D., Graves, A., Antonoglou, I., Wierstra,
  D., and Riedmiller, M.
\newblock Playing atari with deep reinforcement learning.
\newblock \emph{arXiv preprint arXiv:1312.5602}, 2013.

\bibitem[Mnih et~al.(2016)Mnih, Badia, Mirza, Graves, Lillicrap, Harley,
  Silver, and Kavukcuoglu]{mnih2016asynchronous}
Mnih, V., Badia, A.~P., Mirza, M., Graves, A., Lillicrap, T., Harley, T.,
  Silver, D., and Kavukcuoglu, K.
\newblock Asynchronous methods for deep reinforcement learning.
\newblock In \emph{International conference on machine learning}, pp.\
  1928--1937. PMLR, 2016.

\bibitem[Nilsson \& Cully(2021)Nilsson and Cully]{Nilsson2021}
Nilsson, O. and Cully, A.
\newblock Policy gradient assisted map-elites.
\newblock GECCO '21, pp.\  866–875. Association for Computing Machinery,
  2021.

\bibitem[Ota et~al.(2021)Ota, Jha, and Kanezaki]{ota2021training}
Ota, K., Jha, D.~K., and Kanezaki, A.
\newblock Training larger networks for deep reinforcement learning.
\newblock \emph{arXiv preprint arXiv:2102.07920}, 2021.

\bibitem[Parker-Holder et~al.(2020)Parker-Holder, Pacchiano, Choromanski, and
  Roberts]{parker2020effective}
Parker-Holder, J., Pacchiano, A., Choromanski, K.~M., and Roberts, S.~J.
\newblock Effective diversity in population based reinforcement learning.
\newblock \emph{Advances in Neural Information Processing Systems},
  33:\penalty0 18050--18062, 2020.

\bibitem[Paszke et~al.(2019)Paszke, Gross, Massa, Lerer, Bradbury, Chanan,
  Killeen, Lin, Gimelshein, Antiga, Desmaison, Kopf, Yang, DeVito, Raison,
  Tejani, Chilamkurthy, Steiner, Fang, Bai, and Chintala]{NEURIPS2019_9015}
Paszke, A., Gross, S., Massa, F., Lerer, A., Bradbury, J., Chanan, G., Killeen,
  T., Lin, Z., Gimelshein, N., Antiga, L., Desmaison, A., Kopf, A., Yang, E.,
  DeVito, Z., Raison, M., Tejani, A., Chilamkurthy, S., Steiner, B., Fang, L.,
  Bai, J., and Chintala, S.
\newblock Pytorch: An imperative style, high-performance deep learning library.
\newblock In \emph{Advances in Neural Information Processing Systems 32}, pp.\
  8024--8035. 2019.

\bibitem[Petrenko et~al.(2020)Petrenko, Huang, Kumar, Sukhatme, and
  Koltun]{petrenko2020sample}
Petrenko, A., Huang, Z., Kumar, T., Sukhatme, G., and Koltun, V.
\newblock Sample factory: Egocentric 3d control from pixels at 100000 fps with
  asynchronous reinforcement learning.
\newblock In \emph{International Conference on Machine Learning}, pp.\
  7652--7662. PMLR, 2020.

\bibitem[Pierrot et~al.(2022)Pierrot, Mac{\'e}, Chalumeau, Flajolet, Cideron,
  Beguir, Cully, Sigaud, and Perrin]{pierrot2022diversity}
Pierrot, T., Mac{\'e}, V., Chalumeau, F., Flajolet, A., Cideron, G., Beguir,
  K., Cully, A., Sigaud, O., and Perrin, N.
\newblock Diversity policy gradient for sample efficient quality-diversity
  optimization.
\newblock 2022.

\bibitem[Pourchot \& Sigaud(2019)Pourchot and Sigaud]{pourchot2018cem}
Pourchot, A. and Sigaud, O.
\newblock Cem-rl: Combining evolutionary and gradient-based methods for policy
  search.
\newblock In \emph{International Conference on Learning Representations}, 2019.

\bibitem[Raffin et~al.(2021)Raffin, Hill, Gleave, Kanervisto, Ernestus, and
  Dormann]{stable-baselines3}
Raffin, A., Hill, A., Gleave, A., Kanervisto, A., Ernestus, M., and Dormann, N.
\newblock Stable-baselines3: Reliable reinforcement learning implementations.
\newblock \emph{Journal of Machine Learning Research}, 22\penalty0
  (268):\penalty0 1--8, 2021.
\newblock URL \url{http://jmlr.org/papers/v22/20-1364.html}.

\bibitem[Sharma et~al.(2020)Sharma, Gu, Levine, Kumar, and
  Hausman]{sharma2019dynamics}
Sharma, A., Gu, S., Levine, S., Kumar, V., and Hausman, K.
\newblock Dynamics-aware unsupervised discovery of skills.
\newblock In \emph{International Conference on Learning Representations}, 2020.

\bibitem[Stooke \& Abbeel(2018)Stooke and Abbeel]{stooke2018accelerated}
Stooke, A. and Abbeel, P.
\newblock Accelerated methods for deep reinforcement learning.
\newblock \emph{arXiv preprint arXiv:1803.02811}, 2018.

\bibitem[Stooke et~al.(2021)Stooke, Lee, Abbeel, and
  Laskin]{stooke2021decoupling}
Stooke, A., Lee, K., Abbeel, P., and Laskin, M.
\newblock Decoupling representation learning from reinforcement learning.
\newblock In \emph{International Conference on Machine Learning}, pp.\
  9870--9879. PMLR, 2021.

\bibitem[Todorov et~al.(2012)Todorov, Erez, and Tassa]{todorov2012MuJoCo}
Todorov, E., Erez, T., and Tassa, Y.
\newblock Mujoco: A physics engine for model-based control.
\newblock In \emph{2012 IEEE/RSJ International Conference on Intelligent Robots
  and Systems}, pp.\  5026--5033. IEEE, 2012.

\bibitem[Vinyals et~al.(2019)Vinyals, Babuschkin, Czarnecki, Mathieu, Dudzik,
  Chung, Choi, Powell, Ewalds, Georgiev, et~al.]{vinyals2019grandmaster}
Vinyals, O., Babuschkin, I., Czarnecki, W.~M., Mathieu, M., Dudzik, A., Chung,
  J., Choi, D.~H., Powell, R., Ewalds, T., Georgiev, P., et~al.
\newblock Grandmaster level in starcraft ii using multi-agent reinforcement
  learning.
\newblock \emph{Nature}, 575\penalty0 (7782):\penalty0 350--354, 2019.

\end{thebibliography}
\bibliographystyle{icml2022}

\newpage
\appendix
\onecolumn

\begin{table}[ht!]
\caption{Averaged posted price per hour over three main cloud computing platforms for all accelerators considered in this work.}
\label{table-posted-price}
\vskip 0.15in
\begin{center}
\begin{small}
\begin{sc}
\begin{tabular}{lcr}
\toprule
accelerator & price (in dollars per hour)  \\
\midrule
K80    & 0.45  \\
T4     & 0.34   \\
V100    & 2.61  \\
A100    & 2.98  \\
one core (Intel Xeon 2.80Ghz processor) with 2GB of RAM & 0.062  \\
\bottomrule
\end{tabular}
\end{sc}
\end{small}
\end{center}
\vskip -0.1in
\end{table}

\section{Collection of environment interactions}
\label{sec:data-collection}

The studies we have conducted rely on the assumption that training data is available and already loaded in the hardware accelerator memory without delay whenever an update step (or a sequence of 50 update steps for implementations that compile that many update steps in a single execution call) has just completed. We detail how this can be achieved with existing libraries and a few additional coding tools and use these tools in the case studies in Section \ref{sec-case-studies}.

For collecting data, we use a similar distributed setting as in \cite{stooke2018accelerated} for collecting environment interactions. Specifically we use multiple CPU cores and each core is responsible for collecting  the environment interactions for multiple agents in the population. We use the built-in Python \emph{multiprocessing} library for spinning up processes. Agents run independently in parallel, each interacting with its own copy of the environment. Every time new parameters are available (i.e. every 50 update steps in the most efficient implementation considered in this paper), they are copied to host memory and then to shared memory where they can be accessed by all agents which enables us to update the agents in a non-blocking fashion while the main process waits for the next update step execution trace on the hardware accelerator to complete. Note that, based on Table \ref{table-env-step-time} and Figure \ref{fig:shared-critic-runtime}, even when using the fastest hardware accelerator available, the number of CPU cores required to generate enough environment data to take update steps without waiting for a population of up to 80 TD3 agents (while maintaining a ratio of environment steps per update step close to one as in previous studies) never requires more than 30 cpu cores. Note that this is much smaller than the population size because, as the population size increases, the update step time increases while the sampling time per agent remains constant. In practice we use 40 cpu cores to account for overheads due to interprocess-communication of training data between the processes in charge of collecting the data and the processes in charge of carrying out the update steps. Similar observations can be made for SAC agents and other environments.

The environment interaction data is stored in a simple implementation of a replay buffer hosted in the process handling the hardware accelerator(s). The agents add environment data they collect to queues (implemented as \emph{Queue}s from the \emph{multiprocessing} library) and this data is fetched and added to the replay buffer in the background using multiple threads in the process handling the hardware accelerator(s). A simple mechanism is implemented in the replay buffer to block sampling calls (if needed) to guarantee that the update steps per environment step ratio remains close to the target (1 in state-of-the-art implementations). Conversely, the agents are blocked when the queues they populate have reached a maximum size, which happens if the process handling the hardware accelerator(s) is lagging behind in terms of number of update steps performed compared to the total number of environment steps carried out by the agents. We use one replay buffer per agent if data should not be mixed between agents (such as during hyperparameter tuning) or a single one otherwise, such as in \cemrl, \dvd, and \qdpg.

For sampling training data, we spin up an additional background thread in the process handling the hardware accelerator(s) to pre-fetch from the replay buffer(s) as many batches of data as needed to carry out the next update step. Batches of training data for all the agents are concatenated along a new first axis and pre-loaded in the hardware accelerator memory in this background thread. This is done concurrently to waiting for the hardware accelerator to complete the last execution trace.

\begin{table}[ht!]
\caption{Runtime (in milliseconds) for a single interaction with the environment for various MuJoCo Gym environments on an Intel Xeon 2.80Ghz processor with a \jax JIT-compiled policy network.}
\label{table-env-step-time}
\vskip 0.15in
\begin{center}
\begin{small}
\begin{sc}
\begin{tabular}{lccr}
\toprule
 & \tddd & \sac \\
\midrule
HalfCheetah-v2 & 0.94 $\pm$ 0.03 & 0.66 $\pm$ 0.04 \\
Swimmer-v3     & 0.73 $\pm$ 0.08 & 0.65 $\pm$ 0.07 \\
Walker2d-v2    & 1.31 $\pm$ 0.07 & 0.95 $\pm$ 0.09\\
Humanoid-v2    & 1.48 $\pm$ 0.22 & 1.12 $\pm$ 0.19 \\
Hopper-v2      & 1.21 $\pm$ 0.1 & 0.89 $\pm$ 0.11 \\
Ant-v2         & 1.25 $\pm$ 0.15 & 1.21 $\pm$ 0.11 \\
\bottomrule
\end{tabular}
\end{sc}
\end{small}
\end{center}
\vskip -0.1in
\end{table}

\section{Additional experimental results}

\begin{table}[ht!]
\caption{Initial Compilation time (in seconds) for a population of 20 agents for various hardware accelerators on the Halfcheetah-v2 environment and the \emph{Jax (Vectorized)} implementation (50 update steps at once).}
\label{table-compilation-time}
\vskip 0.15in
\begin{center}
\begin{small}
\begin{sc}
\begin{tabular}{lccr}
\toprule
 & TD3 & SAC \\
\midrule
K80   & 7.26 $\pm$ 0.12 & 9.49 $\pm$ 0.15 \\
T4    & 5.32 $\pm$ 0.05 & 7.25 $\pm$ 0.08 \\
V100  & 4.88 $\pm$ 0.07 & 6.80 $\pm$ 0.15 \\
A100  & 4.83 $\pm$ 0.09 & 6.51 $\pm$ 0.05 \\
\bottomrule
\end{tabular}
\end{sc}
\end{small}
\end{center}
\vskip -0.1in
\end{table}

\subsection{\pbt experiments}
\label{sec-experimental-details-pbt}

For this set of experiments, we use 4 T4 accelerators and 40 cpu cores of an Intel Xeon 2.80Ghz processor. The entire population evolves at intervals of 100,000 update steps, the 30\% bottom performing (based on the last 10 episode returns) agents are replaced by copies of randomly sampled agents among the top 30 \% and have their hyperparameters re-sampled from a prior distribution.

For \tddd, we optimize: (1) the learning rates for the policy parameters and the critic parameters with log-uniform distributions with lower bounds 3e-5 and upper bounds 3e-3 so that the distributions are roughly centered around the default values used in state-of-the-art implementations (3e-4), (2) the frequency with which the policy parameters are updated w.r.t. the critic parameters (sampling a uniform value between 0.2 and 1), the (3) noise parameters (sampling uniform values between 0 and 1), as well as the (4) discount factor (sampling a uniform value between 0.9 and 1). All other parameters are kept identical to the original state-of-the-art implementation of \cite{hoffman2020acme}.

For \sac, we optimize: (1) the learning rates for the policy parameters, the critic parameters, and the alpha entropy parameters with log-uniform distributions with lower bounds 3e-5 and upper bounds 3e-3 so that the distributions are roughly centered around the default values in state-of-the-art implementations (3e-4), (2) the target entropy (sampling a uniform value between 0.2 and 2 times the default value used in state-of-the-art-implementations), (3) the reward scale (sampling uniform values between 0.1 and 10), as well as (4) the discount factor (sampling a uniform value between 0.9 and 1). All other parameters are kept identical to the original state-of-the-art implementation of \cite{hoffman2020acme}.

\begin{figure}[ht!]
    \centering
    \includegraphics[width=\textwidth]{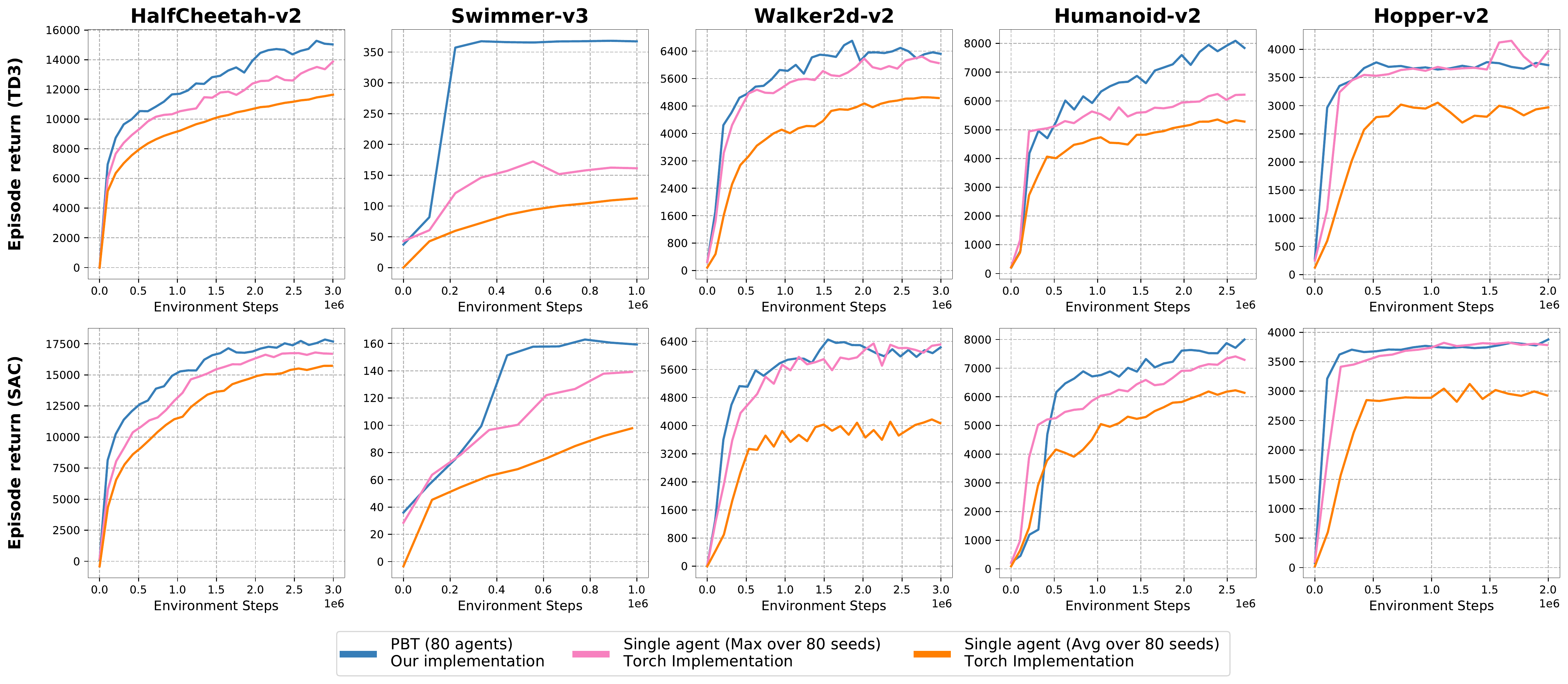}
    \caption{Results for the \pbt implementations where performance is plotted w.r.t the number of timesteps collected in the environment.}
    \label{fig:pbt_steps}
\end{figure}

\subsection{\cemrl and \dvd experiments}

For these sets of experiments, we use one T4 accelerator and 16 cpu cores of an Intel Xeon 2.80Ghz processor. We replicate the studies from \cite{pourchot2018cem} and \cite{parker2020effective} as closely as possible setting aside the second-order change in the order of updates between the critic and the policy parameters mentioned in Section \ref{sec:training-shared-params}. The only differences lie: (1) in the use of the \tddd implementation from the  \acme library \cite{hoffman2020acme} while leaving their hyperparameters untouched, (2) in the use of a schedule for \dvd to update the diversity loss multiplicative factor (whereas the authors use a more complex Multi-Armed-Bandit approach to update it) and (3) an increased initial random noise in \cem from 1e-3 to 1e-2.

\begin{figure}[ht!]
    \centering
    \includegraphics[width=0.95\textwidth]{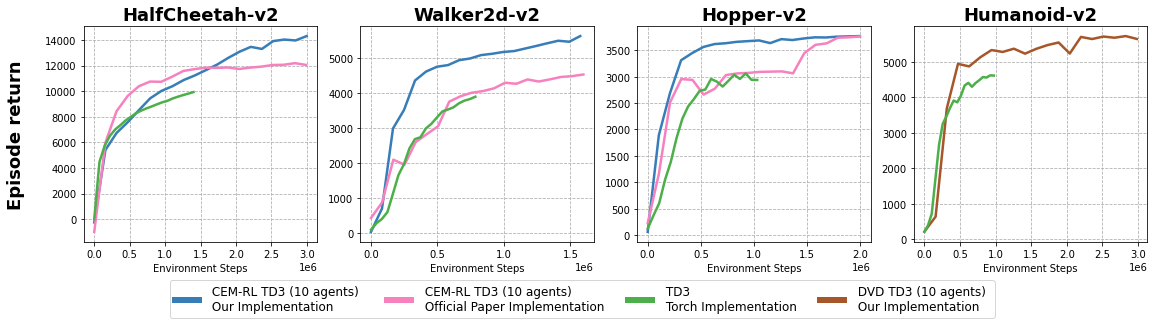}
    \caption{Results for the \cemrl and \dvd implementations where performance is shown against the number of timesteps collected in the environment.}
    \label{fig:cemrl_dvd_steps}
\end{figure}

\newpage

\section{Example of manual vectorization with \torch}
\label{sec-example-vectorization-code}
 The code snippet is for a MultiLayer Perceptron (\mlp) model (which is the core model used by default in the state-of-the-art implementations of \sac and \tddd).
 
\begin{minted}[
fontsize=\footnotesize
]
{python}
from typing import Tuple
import math
import torch


class VectorizedLinearLayer(torch.nn.Module):
    """Vectorized version of torch.nn.Linear."""

    def __init__(
        self,
        population_size: int,
        in_features: int,
        out_features: int,
    ):
        super().__init__()
        self._population_size = population_size
        self._in_features = in_features
        self._out_features = out_features

        self.weight = torch.nn.Parameter(
            torch.empty(self._population_size, self._in_features, self._out_features),
            requires_grad=True,
        )
        self.bias = torch.nn.Parameter(
            torch.empty(self._population_size, 1, self._out_features),
            requires_grad=True,
        )

        for member_id in range(population_size):
            torch.nn.init.kaiming_uniform_(self.weight[member_id], a=math.sqrt(5))
        fan_in, _ = torch.nn.init._calculate_fan_in_and_fan_out(self.weight[0])
        bound = 1 / math.sqrt(fan_in) if fan_in > 0 else 0
        torch.nn.init.uniform_(self.bias, -bound, bound)

    def forward(self, x: torch.Tensor) -> torch.Tensor:
        assert x.shape[0] == self._population_size
        return x.matmul(self.weight) + self.bias


class VectorizedMLP(torch.nn.Module):
    """Vectorized version of MLP."""

    def __init__(
        self,
        population_size: int,
        input_dim: int,
        output_dim: int,
        layers_dim: Tuple[int],
    ):
        super().__init__()

        num_neurons = [input_dim] + list(layers_dim) + [output_dim]
        num_neurons = zip(num_neurons[:-1], num_neurons[1:])

        all_layers = []
        for (in_dim, out_dim) in num_neurons:
            all_layers.append(
                VectorizedLinearLayer(
                    population_size,
                    in_dim,
                    out_dim,
                )
            )
            all_layers.append(torch.nn.ReLU())

        self._mlp = torch.nn.Sequential(*all_layers)
        self._population_size = population_size

    def forward(self, x: torch.Tensor) -> torch.Tensor:
        assert x.shape[0] == self._population_size
        return self._mlp(x)

\end{minted}

\end{document}